\documentclass[]{fairmeta}

\usepackage{amsmath}
\usepackage{float}

\title{CORAL: An LLM-Native Harness for Production Recommender Systems}

\author[1]{Muhammad Rafay Azhar}
\author[1]{Yuhang Zhou}
\author[1]{Gilbert Jiang}
\author[1]{Yuchen Wang}
\author[1]{Rahul Sharma}
\author[1]{Matthew DeSousa}
\author[1]{Jiayi Liu}
\author[1]{Xin Guo}
\author[1]{Lizhu Zhang}
\author[1]{Xiangjun Fan}

\affiliation[1]{Meta AI}

\abstract{Production recommender systems shape what billions of people see, and sustaining their performance is a continual optimization problem: as content, user behavior, and upstream models shift, the choices that govern how these systems retrieve, rank, and serve content must be revisited to keep them near their best operating point. This work has traditionally fallen to human engineers testing changes through online experiments---a slow, reactive process bounded by engineering effort rather than the size of the opportunity, so parts of the system go unrevised and drift as conditions change. Although large language models have been applied to ranking, user modeling, and offline model development, few systems place an agent in a continual, closed loop that acts on a live recommender and learns from the measured consequences of its own decisions. We present CORAL (Constraint-Optimized Recommender via an Agentic Loop), an LLM-native harness that closes such a loop: each cycle, the agent observes operating signals, reasons over a memory of its past decisions and their measured effects, and invokes tools---including a numerical optimizer that keeps every change within a fixed operating budget---to reconfigure the live recommender, after which the measured outcome informs the next cycle. We formulate this as a partially observed, non-stationary, constrained optimization problem in which the policy improves in context, without parameter updates, from the effects of its own prior actions. Across two large-scale social platforms, evaluated with A/B experiments, the same harness improves engagement at no additional serving cost on one and delivers substantial efficiency savings without degrading engagement on the other, together spanning the engagement--efficiency frontier. Its decisions improve as the loop iterates, indicating that a single agentic loop can take on continual optimization work traditionally performed by human algorithm engineers, with a path from human-supervised operation toward autonomous operation under guardrails.}

\date{\today}
\correspondence{Muhammad Rafay Azhar, Yuhang Zhou at \email{\{rafayazhar, zyhang\}@meta.com}\\} 

\begin{document}

\maketitle

\section{Introduction}

Recommender systems mediate how billions of users discover content and products across social media, e-commerce, and entertainment. Over the past decade they have advanced from collaborative filtering to deep learning ranking models~\citep{naumov2019dlrm} and, more recently, to large sequential and generative architectures~\citep{kang2018sasrec, zhai2024hstu}. A modern industrial recommender is not a single model but a large, multi-stage system---candidate retrieval, ranking, and serving---and, at every stage, its pipelines and models are governed by a broad collection of parameters and policies: retrieval budgets, ranking weights, serving and caching policies, and per-segment treatments, among others. Improving such a system is therefore a continual, iterative endeavor: practitioners form hypotheses, implement changes, evaluate them with online experiments, and act on the results, repeating this cycle to advance the system. Effective as it is, this process is bounded by human effort---progress scales with the number of engineers and experiments rather than with the size of the opportunity.

Many of these choices are made by engineers rather than learned end-to-end with the models, and need not remain optimal as the system evolves. The human-driven iteration used to refine them has structural limitations. Each experiment probes only a small region of a vast design space, so improvements are slow and conservative. Because changes are usually prompted by observed regressions or opportunities, the process is reactive rather than anticipatory. And because each cycle is time-intensive and computationally costly, parts of the system are revisited infrequently and largely held fixed between updates---even as content, user behavior, and upstream models continue to shift---so a system can drift away from its own best operating point between interventions. These effects are most pronounced for low-signal and new users, whose behavior is underrepresented in the aggregate metrics that guide most decisions, and they are compounded by operating constraints---serving capacity and compute budgets---that couple decisions across the system.

Large language models (LLMs) offer a way to ease this human bottleneck. Beyond ranking items directly~\citep{hou2024llmrank, zhang2023instructrec}, recent work equips LLMs with memory and tools to reason about users and content~\citep{shen2026mars, chen2026memrec, peng2025survey} and to automate portions of the model-development and system-optimization process~\citep{agentx2026, liu2026nova, hu2026rethinking}. Together, these results suggest that LLM agents can take on work that has traditionally required human algorithm engineers. Most such systems, however, act on the model, the user representation, or the offline development pipeline; comparatively few place an agent in a continual, closed loop that acts on a live production system and learns from the measured consequences of its own decisions.

We present CORAL (Constraint-Optimized Recommender via an Agentic Loop), a flexible, LLM-native agentic harness that closes such a loop around a recommender. In each cycle, the agent observes the system's operating signals, reasons over them together with a memory of its own past decisions and their measured effects, and invokes tools---including numerical optimization---to produce a change; once that change has been deployed and measured, the outcome informs the next cycle. In this way, CORAL continually and autonomously drives improvements that would otherwise demand repeated manual iteration, refining its own decisions over time. We instantiate the harness on a high-value class of decisions: the continuous, constraint-aware allocation of a recommender's resources across its components, where the agent reallocates bounded budgets to improve engagement while respecting operating constraints---treating efficiency as a first-class objective alongside engagement.

We evaluate CORAL on two large-scale social platform surfaces with
different components and decision types, using A/B experiments. On one, it improves engagement---including for low-signal and new users---at no additional serving cost; on the other, it delivers substantial efficiency savings without degrading engagement. The same harness applies to both surfaces, and its decisions improve as the loop iterates.

Our contributions are as follows:
\begin{itemize}
  \item We formulate the continual, agent-driven optimization of a recommender as a partially observed, non-stationary, constrained problem, in which an LLM policy refines its decisions from the measured effects of its own prior actions.
  \item We present the CORAL harness---the context, tools, and closed loop that turn a general-purpose LLM into a continual optimizer of a live recommender---and show that it generalizes across surfaces and decision types rather than being tied to a single lever.
  \item We report two large-scale deployments, evaluated with A/B experiments, that together span the engagement--efficiency frontier, including gains for low-signal and new users.
  \item We distill practical lessons from operating such a loop, including the path from human-supervised operation toward increasingly autonomous operation under guardrails.
\end{itemize}

\section{Related Work}

Our work lies at the intersection of research on tool use and memory in agents and on LLM-based recommender agents~\citep{hou2024llmrank, zhang2023instructrec, zhong2024memorybank}.

\subsection{Tool Use and Memory in LLM Agents}

A broad line of research augments LLM agents with external tools, enabling them to retrieve information, invoke specialized models and APIs~\citep{qin2024toolllm, zhou2026llm, wu2026remember}, execute code~\citep{yang2024swe}, and interact with dynamic environments~\citep{zhao2024toolrec}. Rather than relying solely on parametric knowledge, these agents interleave reasoning, tool selection, execution, and observation to solve tasks requiring external information or specialized capabilities~\citep{feng2025retool}. Subsequent work has advanced tool discovery, multi-tool planning, API-call generation, and generalization to previously unseen tools~\citep{he2025gentool, wu2026swe}. Nevertheless, reliable tool selection, accurate argument construction, and robust recovery from execution failures remain persistent challenges.

Another line of research equips LLM agents with memory mechanisms that preserve information across interactions and support learning from accumulated experience~\citep{chhikara2025mem0, packer2023memgpt, li2025memos, park2023generative}. Existing systems store and retrieve conversational histories, environmental observations, successful strategies, failures, and self-reflections to inform future planning and decision-making~\citep{zhong2024memorybank, ye2026h}. More advanced approaches organize memories into episodic, semantic, reflective, or hierarchical structures and incorporate mechanisms for summarization, consolidation, updating, and selective forgetting~\citep{sumers2023cognitive, xu2026mem}. Despite this progress, fundamental questions remain regarding what information should be retained, how memories should evolve as new evidence arrives, and when outdated, redundant, or conflicting information should be revised or removed.

\subsection{LLM Agents in Recommender Systems}

One line of work uses LLM agents to simulate user behavior for system evaluation and behavioral analysis, contributing primarily to evaluation methodology rather than directly improving recommendation quality~\citep{zhang2024agent4rec, wang2023recagent, zhong2025ggbond, shi2025personax, bougie2025simuser, chen2025recusersim}. A second line develops autonomous agents that reason, plan, and invoke conventional recommender models as tools, but typically handles each interaction independently without maintaining persistent user memory~\citep{wang2024recmind, huang2023interecagent, lei2024macrec, zhao2024toolrec, deng2025recbot, ou2026deepresearch}. Another line equips recommender agents with persistent memory through iterative profile refinement, structured memory organization, and collaborative preference propagation~\citep{xu2025iagent, zhang2024agentcf, liao2026steam, nguyen2026amem4rec, li2026recnet, chen2026memrec}. However, these approaches generally rely on flat memory representations and lack a complete lifecycle governing preference extraction, consolidation, updating, and forgetting.

A more recent line of research shifts the role of agents from generating recommendations to improving recommender systems themselves. These approaches use agents to search over models, code, and system configurations, thereby automating parts of system development and optimization~\citep{agentx2026, liu2026nova, hu2026rethinking, shen2026mars}. In most cases, however, the agent operates on the recommendation model, user representation, or offline development pipeline. Comparatively little attention has been given to agents that continually intervene in a deployed recommender system, observe the measurable effects of their actions, and adapt subsequent decisions based on this feedback. Our work addresses this gap by placing the agent in a continual, closed-loop interaction with a recommender system.

\section{Problem Formulation}

We consider a recommender system whose behavior is governed by a set of tunable control parameters, and we study an agent that updates these parameters over time to improve the system.

Let $s$ denote a configuration of these parameters. Its effect depends
on the surrounding operating context---the users and content the
recommender serves and the upstream models it uses---which the agent
cannot observe directly and which changes over time. We capture this
dependence by indexing the system's response by the cycle $t$: under
configuration $s$ at cycle $t$, the recommender attains a business
objective $J_t(s)$, which we take to be engagement, and incurs an
operating cost $c_t(s)$ that must not exceed a fixed budget $B$. Both
the objective and the operating constraint are set by the operator
rather than by the method; the formulation assumes nothing particular
about either, and any objective that an operator wishes to improve
under a bounded resource constraint fits the same form. It therefore
covers not only maximizing the objective under a cost budget but also
its dual---minimizing cost while holding the objective at a target
level---and an operator may pose either; our two deployments take one
form each.

We group the tunable surface into $N$ control units, indexed $i = 1, \dots, N$; each unit is a component whose setting $s_{t,i}$ the agent controls, drawn from a feasible set---a bounded continuous value or a choice from a discrete set. For example, a control unit might be a retrieval source whose setting is its share of a fixed candidate budget, or a user segment whose setting is one treatment chosen from a discrete serving menu. In each decision cycle $t$, the agent selects a configuration $s_t = (s_{t,1}, \dots, s_{t,N})$.

The best configuration at cycle $t$ is
\begin{equation}
  s^*_t = \arg\max_{s} \; J_t(s) \quad \text{subject to} \quad c_t(s) \le B.
\end{equation}

Here $s^*_t$ is the best feasible configuration at cycle $t$ (the oracle optimum), while $s_t$ is the configuration chosen by the agent. Because the operating context evolves, $s^*_t$ is not fixed: it moves from cycle to cycle, and a configuration left unchanged grows increasingly suboptimal. The agent's aim is therefore not to converge to a single optimum but to keep the deployed configuration close to this moving target while respecting the budget. Across successive cycles, we state this as minimizing the cumulative shortfall relative to the best feasible configuration,
\begin{equation}
  \min_{\pi} \; \sum_{t} \big[\, J_t(s^*_t) - J_t(s_t) \,\big] \quad \text{subject to} \quad c_t(s_t) \le B \;\; \forall t,
\end{equation}
where the configurations $s_t$ are produced by the agent's policy $\pi$.

The policy $\pi$ is an LLM that maps the current observation and memory to the next configuration, $s_t = \pi(o_t, M_t)$. The agent observes the system only through aggregate signals. At each cycle it receives an observation $o_t$ that summarizes per-unit operating statistics over the preceding window, together with their change from the previous window. It also maintains a memory $M_t$ over the most recent $m$ cycles: the observations it has seen, the configurations it has chosen, and the outcomes it has attributed to them. The policy reaches a configuration through a short sequence of actions rather than a single step: it analyzes the current statistics, retrieves relevant history from memory, estimates the effect of its previous configuration, and invokes a numerical optimizer that projects a candidate configuration onto the budget-feasible set, so that every emitted configuration satisfies $c_t(s_t) \le B$ by construction.

\begin{figure}[t]
    \centering
    \includegraphics[width=\textwidth]{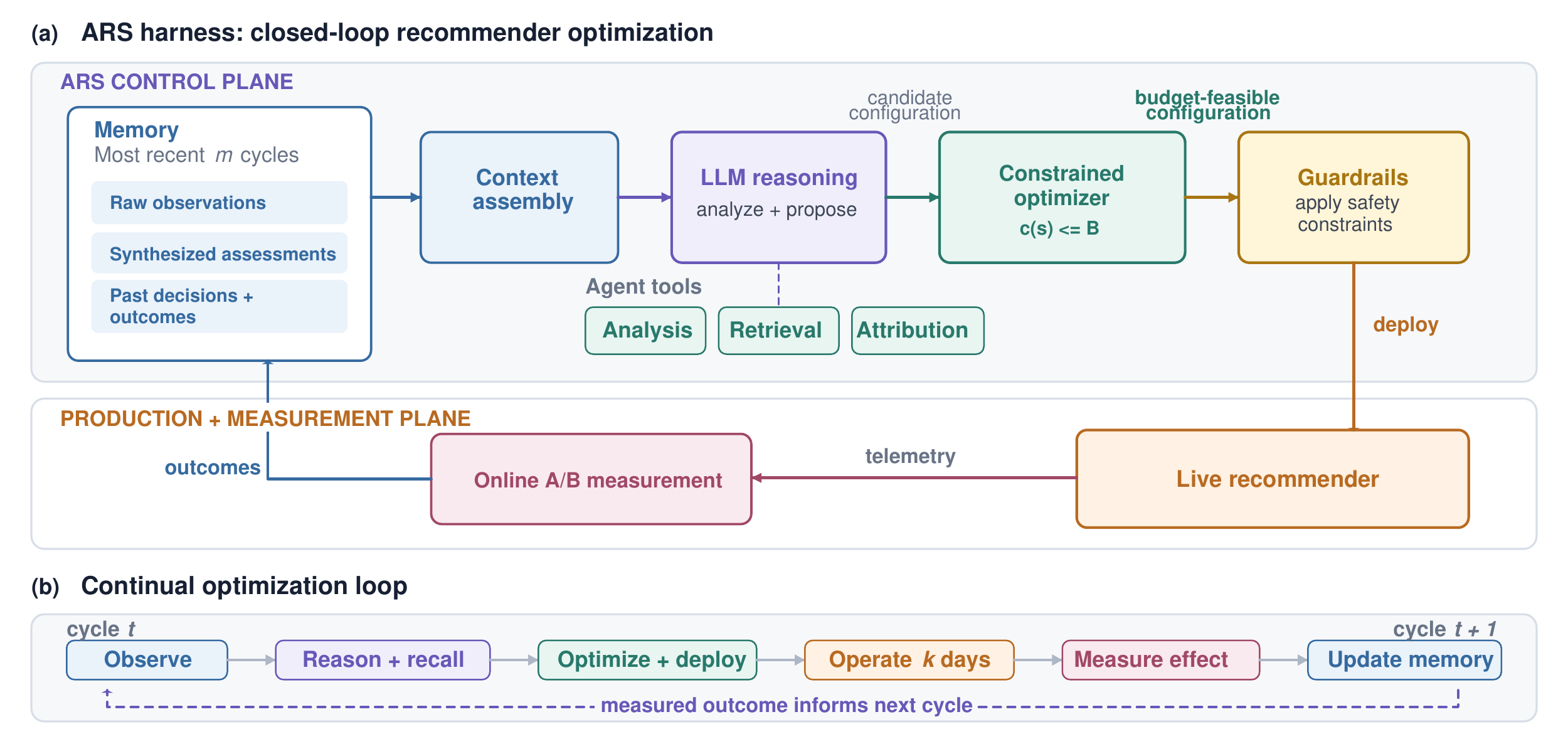}
    \caption{Overview of the CORAL harness. (a) The control plane combines persistent memory and
    LLM reasoning with deterministic tools, constrained optimization, and guardrails. A
    validated, budget-feasible configuration is applied to the live recommender; telemetry and
    online A/B results are measured, attributed, and returned to memory. (b) On each $k$-day
    cycle, the measured effect of the deployed configuration becomes context for the next
    decision, allowing the policy to adapt in context.}
    \label{fig:harness}
  \end{figure}

\section{The CORAL Harness}

The previous section formalizes what the agent must do; here we describe the system that lets a general-purpose language model do it reliably. We call this system the \emph{harness}. It surrounds the language model with three elements the model alone does not provide: the \emph{context} needed to understand the current state of the recommender, a set of \emph{tools} with which to analyze that state and act on it, and a \emph{loop} that runs on a fixed cadence and carries experience from one cycle to the next. The model contributes reasoning; the harness makes that reasoning grounded, budget-feasible, and cumulative. Figure~\ref{fig:harness} shows the overall design.

This problem calls for a policy of an unusual kind. The decision at each
cycle rests on many heterogeneous, partly qualitative per-unit
signals---raw metrics, funnel behavior, and how each unit is
trending---weighed together with domain knowledge about what those
signals imply, and it resists a fixed rule or a single tunable formula.
The right allocation also drifts as the operating context changes, so
the policy must reinterpret the current signals each cycle rather than
settle on a static mapping. An LLM is well suited to both: it reasons
over heterogeneous evidence and prior knowledge to decide where to
reallocate, adapts in context as new outcomes arrive without retraining,
and articulates a rationale for each change---the property the loop
relies on for human oversight and self-attribution. What it cannot
guarantee on its own---a hard budget and consistent, well-formed
decisions---the harness supplies through the optimizer and guardrails.

\subsection{Memory}

The harness maintains a persistent memory that spans the most recent $m$ cycles and gives the model the context to reason about both the system and its own past behavior. We organize it into three stores. An \emph{observation} store records the raw observations $o_t$---the per-unit operating statistics and their recent changes---so the model can see the current state of each part of the system. An \emph{assessment} store holds the model's own synthesized assessments from earlier cycles: concise, natural-language judgments of which parts of the system are performing well or poorly and how their behavior is trending. Finally, a \emph{decision} store records the configurations the agent has previously deployed together with the outcomes later attributed to them. Together, these stores let the model ground each decision not only in the present state of the system but in the consequences of what it has already tried.

\subsection{Tools}

A language model cannot, on its own, guarantee that its proposals are numerically sound or that they respect a hard budget, so the harness equips it with a small set of tools that it invokes while reasoning.
An analysis tool computes and summarizes statistics and their changes from the raw observations.
A retrieval tool surfaces the relevant history---past configurations, assessments, and outcomes---from memory.
An attribution tool estimates the effect of the agent's previous configuration, as described below.
Most important for reliability is a constrained optimizer: given the bounded per-unit adjustments proposed by the model, it returns the closest configuration that satisfies the operating budget---the projection of the proposal onto the budget-feasible set $\{ s : c_t(s) \le B \}$.
When the proposal already respects the budget, this projection returns it unchanged; it binds only when the proposal would overspend, redistributing across units so that the deployed configuration provably satisfies $c_t(s_t) \le B$.
A final tool applies the accepted configuration to the recommender's control surface.
This division of labor lets the model do what it is best at---weighing many heterogeneous signals and articulating why a change should help---while delegating numerical feasibility to a component that can guarantee it.
Each tool is deterministic and returns a structured result the model reasons over, so the model contributes judgment while the tools supply computation and hard guarantees.

\subsection{The Optimization Loop}

The harness runs these components as a closed loop on a fixed cadence of $k$ days, invoking them in the same fixed order every cycle rather
than leaving control flow to the model. It assembles the current
observations and the relevant memory into the model's context; the
model then analyzes the state, recalls what it has tried, estimates the
effect of its last configuration, and proposes a new one, which the
optimizer renders budget-feasible before it is applied. Once deployed,
the configuration remains in effect until the next cycle, and its
effect is measured with an A/B experiment whose result is
written back to memory---closing the loop and becoming part of the
context for the following cycle.

In our deployments we set the cadence to $k=3$ days---long enough for a
configuration's effect to surface in the metrics, yet short enough for the
agent to act promptly on what it observes---and the memory horizon to $m=3$
cycles, which keeps enough recent outcomes for the agent to learn from without
carrying older results that the drifting operating context may have made less
relevant. We chose these as sensible defaults rather than tuning them; the best
cadence may also vary with seasonality, and systematically selecting $k$ and $m$
is a direction we are actively exploring.

This feedback is what separates the loop from a system that is merely re-run on a schedule. Because each configuration persists for a full cycle, the agent can attribute observed changes to its own most recent decision, comparing the period before the change with the period after and, where an A/B experiment is available, obtaining a measured treatment effect. Recording these attributed outcomes lets the agent reinforce the changes that helped and reverse those that did not, so its decisions improve over successive cycles. This improvement requires no retraining: the agent adapts entirely in context, through the observations, assessments, and outcomes accumulated in memory.

The loop runs autonomously, applying each cycle's configuration to the
live recommender directly. A human supervises its operation, tracking
the effect of its changes and the decisions it makes; as confidence in
the agent grows, this supervision is progressively replaced by the
harness's automated guardrails---feasibility checks, bounded-change
limits, and safety constraints.

\section{Case Studies}

We evaluate CORAL on two large-scale social platforms, each
instantiating the harness described above on a different control problem
and a different service, and each evaluated with A/B experiments. The
first targets engagement and shows how the loop improves across cycles
and benefits low-signal users; the second targets serving efficiency
and shows that the same harness transfers to a very different decision.

\begin{table}[t]
  \centering
  \caption{Effect of CORAL's retrieval-budget policy on a
  large-scale video service across three successive rounds
  (R1--R3) of the loop, each evaluated with an A/B
  experiment. R1 is a zero-shot proposal; R3 aggregates several
  decision cycles and is the deployed configuration. Sessions are
  video-viewing sessions; ``neutral'' denotes no statistically
  significant change.}
  \label{tab:fbv-rounds}
  \begin{tabular}{lccc}
    \toprule
    & R1 & R2 & R3 \\
    \midrule
    Watch time                 & $+0.13\%$ & neutral & $+0.15\%$ \\
    Sessions (all users)       & neutral   & neutral & $+0.16\%$ \\
    Sessions (largest market)  & neutral   & neutral & $+0.77\%$ \\
    \bottomrule
  \end{tabular}
\end{table}

\subsection{Allocating Retrieval Budget Across Candidate Sources}

Our first deployment is a video-recommendation service that assembles each user's candidates from a set of complementary retrieval sources. Each source is allotted a share of a fixed retrieval budget, determining how many candidates it may contribute. These shares are typically hand-set and seldom revised, yet the best allocation shifts as content and behavior change, and a source that is efficient for one population can be wasteful for another.

In our formulation, the control units are the retrieval sources; a configuration assigns each source a budget multiplier within a bounded range, and the operating cost is the total retrieval budget consumed. Each cycle, the agent observes per-source signals---how many items a source contributes, how well those items convert into engaged views, and how far they survive the downstream funnel---and proposes a reallocation, trimming budget from sources whose candidates convert poorly and redirecting it to sources that deliver engaged views efficiently, all within the total budget.

The agent operates as an autonomous closed loop under human
oversight, and its policy improved over three successive rounds of
the loop, each evaluated with an A/B experiment
(Table~\ref{tab:fbv-rounds}). A first, zero-shot proposal, formed
from a single window of statistics, produced a small watch-time
gain but no significant change in sessions (sessions are individual user app visits containing at least
one video view). In the second round
the agent shifted budget more aggressively between sources but
overcorrected, and its measured effect was neutral. Incorporating
that outcome, the agent refined the allocation over several
further cycles and arrived at the deployed configuration reported
below, which improved watch time further and produced significant
session gains. This progression is not monotonic---the second round did not
improve on the first---yet it is precisely the behavior the loop
is designed to produce: rather than a gain in every round, the
agent reacts to the measured effect of its own decisions and, over
successive rounds, converges on a better configuration.

For the converged global allocation, the A/B experiment, spanning
millions of users, showed a 0.16\% increase in video-viewing sessions
across all users, alongside a 0.15\% increase in total watch time---all
at no additional serving cost, as the reallocation consolidated the
retrieval budget rather than expanding it.

We then extended the agent beyond a single global allocation to produce distinct allocations for different user segments, defined by engagement level and account tenure---for example, highly active users, low-signal users, and newly joined users. This per-segment control matters most for low-signal and new users, whom an allocation tuned to highly active users tends to
underserve. For this cohort, whose historical engagement is
sparse, the agent shifted budget toward retrieval sources that
draw on content and current-context signals---which stay reliable
when per-user data is scarce---and away from sources that rely on
rich user histories. This raised video-viewing sessions for new
low-signal users by 0.23\%. The result indicates that the same harness can specialize its policy for segments that a single global allocation leaves behind, directly addressing the systematic under-service of low-signal users.

\begin{table}[t]
  \centering
  \caption{Per-call cost parameters for the two deployments (estimated from prompt and payload sizes; the pipelines do not log token usage).}
  \label{tab:cost}
  \begin{tabular}{lccc}
    \toprule
    Case study & Calls per cycle & Input tokens/call & Output tokens/call \\
    \midrule
    Retrieval-budget & $\sim$10 & $\sim$1{,}500 & $\sim$2{,}500 \\
    Serving-capacity & $\sim 8$ & $\sim$2{,}000 & $\sim$2{,}500 \\
    \bottomrule
  \end{tabular}
\end{table}

\subsection{Allocating Serving Capacity Across User Segments}

Our second experiment is on a different service, where the agent
allocates serving capacity across user segments. For each segment, the
service can select a treatment from a menu that ranges from lightweight
to compute-intensive, varying how aggressively it retrieves and ranks,
how much it caches, and how much it prefetches. More intensive
treatments can raise engagement but cost more to serve, and the total
serving cost is capped by a fixed budget---so choosing a treatment per
segment is precisely the constrained allocation our formulation
describes.

Here the control units are the user segments; a configuration assigns each segment a treatment from the discrete menu, and the operating cost is the compute required to serve it, bounded by the budget. Each cycle, the agent observes per-segment engagement and cost and decides where to spend more and where to spend less---raising the treatment for segments where added compute yields the most engagement and lowering it for segments where it yields little, so that the reclaimed budget funds the increases.

Because a treatment assignment stays in effect for a full cycle, the agent can compare a segment's engagement and cost before and after its previous decision, attribute the change to that decision, and use this to decide whether to continue in that direction or reverse course in the next cycle.

This behavior is evident in our experiments. In an A/B test involving
millions of users, the agent improved its result over two successive
rounds. In the first, working within a subset of user segments, the
agent reduced serving cost substantially, saving millions of USD in
annualized capacity expenditure. Reading that result against its
previous decision, the agent recognized that the same change could be
applied safely to additional user segments, and in the second round it
widened its allocation to include them, increasing the savings of the
first round by 44\% while leaving engagement statistically
unchanged---and thereby freeing capacity that can be reinvested
elsewhere. This deployment exercises the efficiency side of the
objective: it avoids degrading engagement while directing the operating
budget where it is most productive.

\subsection{Discussion}

Across the two studies, the same harness addressed markedly different
control problems---a continuous reallocation of retrieval budget across
sources and a discrete assignment of serving treatments across
segments---and improved the system along complementary axes:
engagement, including for low-signal users, in the first, and serving
efficiency without degrading engagement in the second. Together they trace the engagement--efficiency frontier that a production recommender must manage, and they show that a single, flexible harness generalizes across different services and decision types. These gains came from a process that required no per-decision
engineering effort. Producing such allocations by hand is a heavy,
periodic undertaking: an engineer forms a hypothesis, runs an
experiment, and revises a single lever over the course of weeks, with
the effort growing as the number of levers and segments grows. The
harness instead adjusts every control unit on a short, fixed cadence,
autonomously and at negligible compute cost---compressing a tuning
cycle from several engineer-weeks to a few autonomous days, an
order-of-magnitude faster turnaround with no engineer in the loop.

\section{Computational Cost}
\label{app:cost}

Because the harness acts on the control surface rather than on individual requests, its language-model cost is charged per decision cycle, not per user. Over a deployment spanning $T$ days at a cadence of $k$ days, the total inference cost is
\begin{equation}
\text{Cost} = \underbrace{(T/k)}_{\text{cycles}} \cdot C \cdot \big(\tau_{\text{in}}\,p_{\text{in}} + \tau_{\text{out}}\,p_{\text{out}}\big),
\label{eq:cost}
\end{equation}
where $C$ is the number of LLM calls per cycle, $\tau_{\text{in}}$ and $\tau_{\text{out}}$ are the average input and output tokens per call, and $p_{\text{in}}, p_{\text{out}}$ are the per-token prices. Three properties follow. First, cost is inversely proportional to the cadence: a shorter $k$ improves responsiveness at proportionally higher cost. Second, it is bounded per cycle, since $\tau_{\text{out}}$ cannot exceed the model's output-token limit. Third---and most consequential at scale---$C$ is fixed by how the control surface is partitioned into decision groups (a handful per cycle), not by the number of users or requests served, so the cost is independent of traffic and remains negligible even on billion-user surfaces. This is in stark contrast to per-item or per-user LLM inference, where cost grows with traffic.

Table~\ref{tab:cost} lists per-call parameters for the two deployments. Each cycle issues only a handful of calls of a few thousand tokens each, so over a deployment of several cycles the total is on the order of $10^{6}$ tokens. Using representative frontier-LLM pricing as a reference, this places the end-to-end inference cost of each deployment on the order of tens of U.S. dollars---negligible against the engagement gains and operating-cost savings it produces.

\section{Conclusion}

We presented CORAL, an LLM-native harness that places a language model
in a continual, closed loop around a recommender. Rather than optimizing
a model offline or serving recommendations directly, the agent acts on
the recommender's live control surface: each cycle it observes the
system's operating signals, reasons over them together with a memory of
its past decisions and their measured effects, applies a change through
tools that keep it budget-feasible, and learns from the measured
outcome. We formulated this as a partially observed, non-stationary,
constrained optimization problem, and showed that an LLM policy can
address it by refining its decisions in context, without retraining.
Across two large-scale social platforms, evaluated with A/B experiments,
the same harness improved both engagement---including for low-signal
and new users---and serving efficiency, providing evidence that a single
agentic loop can carry out the continual optimization work that has
traditionally fallen to human algorithm engineers.

Our approach has limitations, several of which point to future work.
First, although the loop runs autonomously, it still operates under
human supervision; strengthening the harness's automated guardrails so
that this supervision can be reduced is a natural next step. Second,
both studies instantiate the harness on a single class of decision---the
constrained allocation of a recommender's resources across its
components---and extending the same loop to qualitatively different
levers, such as the retrieval and ranking logic itself, would further
test the generality it is designed for. Finally, our evidence comes
from A/B experiments that measure real effects but are costly to run and
specific to their setting; a standardized way to evaluate agent-driven
system optimization before deployment remains an open problem, and one
we hope this work helps motivate.

\clearpage
\newpage
\bibliographystyle{assets/plainnat}
\bibliography{coral}

\begin{thebibliography}{43}
\providecommand{\natexlab}[1]{#1}
\providecommand{\url}[1]{\texttt{#1}}
\expandafter\ifx\csname urlstyle\endcsname\relax
  \providecommand{\doi}[1]{doi: #1}\else
  \providecommand{\doi}{doi: \begingroup \urlstyle{rm}\Url}\fi

\bibitem[Bougie and Watanabe(2025)]{bougie2025simuser}
Nicolas Bougie and Narimasa Watanabe.
\newblock Simuser: Simulating user behavior with large language models for recommender system evaluation.
\newblock In \emph{Proceedings of the 63rd Annual Meeting of the Association for Computational Linguistics (Volume 5: Industry Track)}, 2025.

\bibitem[Chen et~al.(2025)Chen, Dai, Zhang, Feng, Zhang, Tang, Chen, Zhu, and Dong]{chen2025recusersim}
Luyu Chen, Quanyu Dai, Zeyu Zhang, Xueyang Feng, Mingyu Zhang, Pengcheng Tang, Xu~Chen, Yue Zhu, and Zhenhua Dong.
\newblock Recusersim: A realistic and diverse user simulator for evaluating conversational recommender systems.
\newblock In \emph{Proceedings of the ACM Web Conference 2025, Industry Track}, 2025.

\bibitem[Chen et~al.(2026)Chen, Zhao, Huang, Ye, Ju, Zhao, Shah, Chen, and Zhang]{chen2026memrec}
Weixin Chen, Yuhan Zhao, Jingyuan Huang, Zihe Ye, Clark~Mingxuan Ju, Tong Zhao, Neil Shah, Li~Chen, and Yongfeng Zhang.
\newblock Memrec: Collaborative memory-augmented agentic recommender system.
\newblock \emph{arXiv preprint arXiv:2601.08816}, 2026.

\bibitem[Chhikara et~al.(2025)Chhikara, Khant, Aryan, Singh, and Yadav]{chhikara2025mem0}
Prateek Chhikara, Dev Khant, Saket Aryan, Taranjeet Singh, and Deshraj Yadav.
\newblock Mem0: Building production-ready ai agents with scalable long-term memory.
\newblock \emph{arXiv preprint arXiv:2504.19413}, 2025.

\bibitem[Deng et~al.(2025)Deng, Lian, Lei, Gao, Huang, and Chen]{deng2025recbot}
Yu~Deng, Jianxun Lian, Yuxuan Lei, Chongming Gao, Kexin Huang, and Jiawei Chen.
\newblock Recbot: Agent-based recommendation system.
\newblock \emph{arXiv preprint arXiv:2509.21317}, 2025.

\bibitem[Feng et~al.(2025)Feng, Huang, Qu, Zhang, Qin, Zhong, Jiang, Chi, and Zhong]{feng2025retool}
Jiazhan Feng, Shijue Huang, Xingwei Qu, Ge~Zhang, Yujia Qin, Baoquan Zhong, Chengquan Jiang, Jinxin Chi, and Wanjun Zhong.
\newblock Retool: Reinforcement learning for strategic tool use in llms.
\newblock \emph{arXiv preprint arXiv:2504.11536}, 2025.

\bibitem[He et~al.(2025)He, Neville, Wan, Yang, Liu, Xu, Song, Pan, and Zhou]{he2025gentool}
Jie He, Jennifer Neville, Mengting Wan, Longqi Yang, Hui Liu, Xiaofeng Xu, Xia Song, Jeff~Z Pan, and Pei Zhou.
\newblock Gentool: Enhancing tool generalization in language models through zero-to-one and weak-to-strong simulation.
\newblock In \emph{Findings of the Association for Computational Linguistics: ACL 2025}, pages 1097--1122, 2025.

\bibitem[Hou et~al.(2024)Hou, Zhang, Lin, Lu, Xie, McAuley, and Zhao]{hou2024llmrank}
Yupeng Hou, Junjie Zhang, Zihan Lin, Hongyu Lu, Ruobing Xie, Julian McAuley, and Wayne~Xin Zhao.
\newblock Large language models are zero-shot rankers for recommender systems.
\newblock In \emph{Proceedings of the 46th European Conference on Information Retrieval (ECIR)}, 2024.

\bibitem[Hu et~al.(2026)Hu, Deng, Mu, Zhang, Wang, Zhang, and Zeng]{hu2026rethinking}
Jinxin Hu, Hao Deng, Lingyu Mu, Hao Zhang, Shizhun Wang, Yu~Zhang, and Xiaoyi Zeng.
\newblock Rethinking recommendation paradigms: From pipelines to agentic recommender systems, 2026.
\newblock \url{https://arxiv.org/abs/2603.26100}.

\bibitem[Huang et~al.(2023)Huang, Lian, Lei, Yao, Lian, and Xie]{huang2023interecagent}
Xu~Huang, Jianxun Lian, Yuxuan Lei, Jing Yao, Defu Lian, and Xing Xie.
\newblock Recommender ai agent: Integrating large language models for interactive recommendations.
\newblock \emph{arXiv preprint arXiv:2308.16505}, 2023.

\bibitem[Kang and McAuley(2018)]{kang2018sasrec}
Wang-Cheng Kang and Julian McAuley.
\newblock Self-attentive sequential recommendation.
\newblock In \emph{Proceedings of the IEEE International Conference on Data Mining (ICDM)}, 2018.

\bibitem[Lao et~al.(2026)Lao, Pan, Ma, Li, Lin, Shi, Zhao, Gai, Zhou, Zhou, Chen, Yang, Bie, Yi, Yang, Yang, Li, Xie, Lv, Wang, Wang, Chen, Huang, Wang, Zhao, Zhuang, Xia, Liu, Ma, He, Cong, Jiang, Wang, Xia, Xu, Xie, Qiao, Liang, Yue, Wang, Yang, Jia, Qin, Wang, Li, Song, Xu, Luo, Tang, Liu, Jin, Wang, Zhang, Gao, Li, Luo, Ning, Liu, Liu, Guo, Liu, and Cui]{agentx2026}
Changxin Lao, Fei Pan, Guozhuang Ma, Han Li, Huihuang Lin, Jijun Shi, Kangzhi Zhao, Kun Gai, Mo~Zhou, Qinqin Zhou, Quan Chen, Ruochen Yang, Shifu Bie, Shijie Yi, Shuang Yang, Shuo Yang, Wenhao Li, Wentao Xie, Xiao Lv, Xuming Wang, Yijun Wang, Yiming Chen, Yusheng Huang, Zhongyuan Wang, Zibo Zhao, Zijie Zhuang, Baoning Xia, Chao Liu, Chaoyi Ma, Chubo He, Dawei Cong, Feng Jiang, Gang Wang, Guilin Xia, Hanwen Xu, Jiahong Xie, Jiahui Qiao, Jian Liang, Jiangfan Yue, Jing Wang, Jinghan Yang, Jinghui Jia, Kan Qin, Lei Wang, Ming Li, Peilin Song, Pengbo Xu, Qiang Luo, Ruiming Tang, Shiyang Liu, Shuxian Jin, Tao Wang, Tao Zhang, Xiang Gao, Xianghan Li, Yingsong Luo, Yiwen Ning, Yongcheng Liu, Yueyang Liu, Yuan Guo, Zhaojie Liu, and Zhenkai Cui.
\newblock Agentx: Towards agent-driven self-iteration of industrial recommender systems, 2026.
\newblock \url{https://arxiv.org/abs/2606.26859}.

\bibitem[Lei et~al.(2024)Lei, Wang, Zhang, and Chen]{lei2024macrec}
Zhefan Lei, Hengxu Wang, Jiawei Zhang, and Shuai Chen.
\newblock Macrec: A multi-agent collaboration framework for recommendation.
\newblock \emph{arXiv preprint arXiv:2402.15235}, 2024.

\bibitem[Li et~al.(2026)Li, Wang, Li, Li, Zhang, Chen, Zhao, and Wen]{li2026recnet}
Bingqian Li, Xiaolei Wang, Junyi Li, Weitao Li, Long Zhang, Sheng Chen, Wayne~Xin Zhao, and Ji-Rong Wen.
\newblock Recnet: Self-evolving preference propagation for agentic recommender systems.
\newblock \emph{arXiv preprint arXiv:2601.21609}, 2026.

\bibitem[Li et~al.(2025)Li, Xi, Li, Chen, Chen, Song, Niu, Wang, Yang, Tang, et~al.]{li2025memos}
Zhiyu Li, Chenyang Xi, Chunyu Li, Ding Chen, Boyu Chen, Shichao Song, Simin Niu, Hanyu Wang, Jiawei Yang, Chen Tang, et~al.
\newblock Memos: A memory os for ai system.
\newblock \emph{arXiv preprint arXiv:2507.03724}, 2025.

\bibitem[Liao et~al.(2026)Liao, Wu, Hou, Wang, Wu, and Wang]{liao2026steam}
Yuxin Liao, Le~Wu, Min Hou, Yu~Wang, Han Wu, and Meng Wang.
\newblock From atom to community: Structured and evolving agent memory for user behavior modeling.
\newblock \emph{arXiv preprint arXiv:2601.16872}, 2026.

\bibitem[Liu et~al.(2026)Liu, Fang, Sun, Huang, Luo, Liu, Chen, Liu, Ma, Chai, Wang, Quan, Cui, Zhu, Chen, Xu, Xiao, Gu, and Jiang]{liu2026nova}
Shaohua Liu, Liang Fang, Yilong Sun, Shudong Huang, Qingsong Luo, Shaoxin Liu, Xiaoyang Chen, Dongqiang Liu, Chuangang Ma, Zhenzhen Chai, Henghuan Wang, Shijie Quan, Changyuan Cui, Zhangbin Zhu, Peng Chen, Wei Xu, Lei Xiao, Haijie Gu, and Jie Jiang.
\newblock Nova: A verification-aware agent harness for architecture evolution in industrial recommender systems, 2026.
\newblock \url{https://arxiv.org/abs/2606.27243}.

\bibitem[Naumov et~al.(2019)Naumov, Mudigere, Shi, Huang, Sundaraman, Park, Wang, Gupta, Wu, Azzolini, Dzhulgakov, Mallevich, Cherniavskii, Lu, Krishnamoorthi, Yu, Kondratenko, Pereira, Chen, Chen, Rao, Jia, Xiong, and Smelyanskiy]{naumov2019dlrm}
Maxim Naumov, Dheevatsa Mudigere, Hao-Jun~Michael Shi, Jianyu Huang, Narayanan Sundaraman, Jongsoo Park, Xiaodong Wang, Udit Gupta, Carole-Jean Wu, Alisson~G. Azzolini, Dmytro Dzhulgakov, Andrey Mallevich, Ilia Cherniavskii, Yinghai Lu, Raghuraman Krishnamoorthi, Ansha Yu, Volodymyr Kondratenko, Stephanie Pereira, Xianjie Chen, Wenlin Chen, Vijay Rao, Bill Jia, Liang Xiong, and Misha Smelyanskiy.
\newblock Deep learning recommendation model for personalization and recommendation systems, 2019.
\newblock \url{https://arxiv.org/abs/1906.00091}.

\bibitem[Nguyen et~al.(2026)Nguyen, Kieu, and Le]{nguyen2026amem4rec}
Minh-Duc Nguyen, Hai-Dang Kieu, and Dung~D. Le.
\newblock Amem4rec: Leveraging cross-user similarity for memory evolution in agentic llm recommenders.
\newblock \emph{arXiv preprint arXiv:2602.08837}, 2026.

\bibitem[Ou et~al.(2026)Ou, Wu, Wang, Zheng, Zhao, Li, Zhang, Chen, and Wen]{ou2026deepresearch}
Kesha Ou, Chenghao Wu, Xiaolei Wang, Bowen Zheng, Wayne~Xin Zhao, Weitao Li, Long Zhang, Sheng Chen, and Ji-Rong Wen.
\newblock Deep research for recommender systems.
\newblock \emph{arXiv preprint arXiv:2603.07605}, 2026.

\bibitem[Packer et~al.(2023)Packer, Wooders, Lin, Fang, Patil, Stoica, and Gonzalez]{packer2023memgpt}
Charles Packer, Sarah Wooders, Kevin Lin, Vivian Fang, Shishir~G. Patil, Ion Stoica, and Joseph~E. Gonzalez.
\newblock Memgpt: Towards llms as operating systems.
\newblock \emph{arXiv preprint arXiv:2310.08560}, 2023.

\bibitem[Park et~al.(2023)Park, O'Brien, Cai, Morris, Liang, and Bernstein]{park2023generative}
Joon~Sung Park, Joseph O'Brien, Carrie~Jun Cai, Meredith~Ringel Morris, Percy Liang, and Michael~S Bernstein.
\newblock Generative agents: Interactive simulacra of human behavior.
\newblock In \emph{Proceedings of the 36th annual acm symposium on user interface software and technology}, pages 1--22, 2023.

\bibitem[Peng et~al.(2025)Peng, Liu, Huang, Yang, and Shao]{peng2025survey}
Qiyao Peng, Hongtao Liu, Hua Huang, Qing Yang, and Minglai Shao.
\newblock A survey on llm-powered agents for recommender systems, 2025.
\newblock \url{https://arxiv.org/abs/2502.10050}.

\bibitem[Qin et~al.(2024)Qin, Liang, Ye, Zhu, Yan, Lu, Lin, Cong, Tang, Qian, et~al.]{qin2024toolllm}
Yujia Qin, Shihao Liang, Yining Ye, Kunlun Zhu, Lan Yan, Yaxi Lu, Yankai Lin, Xin Cong, Xiangru Tang, Bill Qian, et~al.
\newblock Toolllm: Facilitating large language models to master 16000+ real-world apis.
\newblock In \emph{International Conference on Learning Representations}, volume 2024, pages 9695--9717, 2024.

\bibitem[Shen et~al.(2026)Shen, Zhou, Wu, Zhao, Lin, Huang, Zhong, Zhang, Zhang, Fan, and Yan]{shen2026mars}
Xiang Shen, Yuhang Zhou, Yifan Wu, Zhuokai Zhao, Siyu Lin, Lei Huang, Qianqian Zhong, Lizhu Zhang, Benyu Zhang, Xiangjun Fan, and Hong Yan.
\newblock Agentic recommender system with hierarchical belief-state memory, 2026.
\newblock \url{https://arxiv.org/abs/2605.14401}.

\bibitem[Shi et~al.(2025)Shi, Xu, Zhang, Zi, Wu, and Xu]{shi2025personax}
Yunxiao Shi, Wujiang Xu, Zeqi Zhang, Xing Zi, Qiang Wu, and Min Xu.
\newblock Personax: A recommendation agent oriented user modeling framework for long behavior sequence.
\newblock In \emph{Findings of the Association for Computational Linguistics (ACL)}, 2025.

\bibitem[Sumers et~al.(2023)Sumers, Yao, Narasimhan, and Griffiths]{sumers2023cognitive}
Theodore Sumers, Shunyu Yao, Karthik~R Narasimhan, and Thomas~L Griffiths.
\newblock Cognitive architectures for language agents.
\newblock \emph{Transactions on Machine Learning Research}, 2023.

\bibitem[Wang et~al.(2023)Wang, Zhang, Yang, Chen, Tang, Zhang, Chen, Lin, Song, Zhao, Xu, Dou, Wang, and Wen]{wang2023recagent}
Lei Wang, Jingsen Zhang, Hao Yang, Zhiyuan Chen, Jiakai Tang, Zeyu Zhang, Xu~Chen, Yankai Lin, Ruihua Song, Wayne~Xin Zhao, Jun Xu, Zhicheng Dou, Jun Wang, and Ji-Rong Wen.
\newblock User behavior simulation with large language model based agents.
\newblock \emph{arXiv preprint arXiv:2306.02552}, 2023.

\bibitem[Wang et~al.(2024)Wang, Jiang, Chen, Yang, Zhou, Cho, Fan, Lu, Huang, and Lu]{wang2024recmind}
Yancheng Wang, Ziyan Jiang, Zheng Chen, Fan Yang, Yingxue Zhou, Eunah Cho, Xing Fan, Yanbin Lu, Xiaojiang Huang, and Yingbo Lu.
\newblock Recmind: Large language model powered agent for recommendation.
\newblock \emph{arXiv preprint arXiv:2308.14296}, 2024.

\bibitem[Wu et~al.(2026{\natexlab{a}})Wu, Zhang, Zhou, Wang, Peng, Li, Fan, and Zhao]{wu2026remember}
Yifan Wu, Lizhu Zhang, Yuhang Zhou, Mingyi Wang, Bo~Peng, Serena Li, Xiangjun Fan, and Zhuokai Zhao.
\newblock Remember when it matters: Proactive memory agent for long-horizon agents.
\newblock \emph{arXiv preprint arXiv:2607.08716}, 2026{\natexlab{a}}.

\bibitem[Wu et~al.(2026{\natexlab{b}})Wu, Zhao, Li, Lee, Zhu, Wu, Yu, Li, Zhang, Fan, et~al.]{wu2026swe}
Yifan Wu, Zhuokai Zhao, Songlin Li, Ho~Hin Lee, Jiacheng Zhu, Shirley Wu, Tianhe Yu, Serena Li, Lizhu Zhang, Xiangjun Fan, et~al.
\newblock Swe-together: Evaluating coding agents in interactive user sessions.
\newblock \emph{arXiv preprint arXiv:2606.29957}, 2026{\natexlab{b}}.

\bibitem[Xu et~al.(2025)Xu, Shi, Liang, Ning, Mei, Wang, Zhu, Xu, and Zhang]{xu2025iagent}
Wujiang Xu, Yunxiao Shi, Zujie Liang, Xuying Ning, Kai Mei, Kun Wang, Xi~Zhu, Min Xu, and Yongfeng Zhang.
\newblock iagent: Llm agent as a shield between user and recommender systems.
\newblock In \emph{Proceedings of the 63rd Annual Meeting of the Association for Computational Linguistics (ACL)}, 2025.

\bibitem[Xu et~al.(2026)Xu, Liang, Mei, Gao, Tan, and Zhang]{xu2026mem}
Wujiang Xu, Zujie Liang, Kai Mei, Hang Gao, Juntao Tan, and Yongfeng Zhang.
\newblock A-mem: Agentic memory for llm agents.
\newblock \emph{Advances in Neural Information Processing Systems}, 38:\penalty0 17577--17604, 2026.

\bibitem[Yang et~al.(2024)Yang, Jimenez, Wettig, Lieret, Yao, Narasimhan, and Press]{yang2024swe}
John Yang, Carlos Jimenez, Alexander Wettig, Kilian Lieret, Shunyu Yao, Karthik Narasimhan, and Ofir Press.
\newblock Swe-agent: Agent-computer interfaces enable automated software engineering.
\newblock \emph{Advances in Neural Information Processing Systems}, 37:\penalty0 50528--50652, 2024.

\bibitem[Ye et~al.(2026)Ye, Huang, Chen, and Zhang]{ye2026h}
Zihe Ye, Jingyuan Huang, Weixin Chen, and Yongfeng Zhang.
\newblock H-mem: Hybrid multi-dimensional memory management for long-context conversational agents.
\newblock In \emph{Proceedings of the 19th Conference of the European Chapter of the Association for Computational Linguistics (Volume 1: Long Papers)}, pages 7756--7775, 2026.

\bibitem[Zhai et~al.(2024)Zhai, Liao, Liu, Wang, Li, Cao, Gao, Gong, Gu, He, Lu, and Shi]{zhai2024hstu}
Jiaqi Zhai, Lucy Liao, Xing Liu, Yueming Wang, Rui Li, Xuan Cao, Leon Gao, Zhaojie Gong, Fangda Gu, Michael He, Yinghai Lu, and Yu~Shi.
\newblock Actions speak louder than words: Trillion-parameter sequential transducers for generative recommendations, 2024.
\newblock \url{https://arxiv.org/abs/2402.17152}.

\bibitem[Zhang et~al.(2024{\natexlab{a}})Zhang, Chen, Sheng, Wang, and Chua]{zhang2024agent4rec}
An~Zhang, Yuxin Chen, Leheng Sheng, Xiang Wang, and Tat-Seng Chua.
\newblock On generative agents in recommendation.
\newblock In \emph{Proceedings of the 47th International ACM SIGIR Conference on Research and Development in Information Retrieval}, 2024{\natexlab{a}}.

\bibitem[Zhang et~al.(2023)Zhang, Xie, Hou, Zhao, Lin, and Wen]{zhang2023instructrec}
Junjie Zhang, Ruobing Xie, Yupeng Hou, Wayne~Xin Zhao, Leyu Lin, and Ji-Rong Wen.
\newblock Recommendation as instruction following: A large language model empowered recommendation approach, 2023.
\newblock \url{https://arxiv.org/abs/2305.07001}.

\bibitem[Zhang et~al.(2024{\natexlab{b}})Zhang, Hou, Xie, Sun, McAuley, Zhao, Lin, and Wen]{zhang2024agentcf}
Junjie Zhang, Yupeng Hou, Ruobing Xie, Wenqi Sun, Julian McAuley, Wayne~Xin Zhao, Leyu Lin, and Ji-Rong Wen.
\newblock Agentcf: Collaborative learning with autonomous language agents for recommender systems.
\newblock In \emph{Proceedings of the ACM Web Conference (WWW)}, 2024{\natexlab{b}}.

\bibitem[Zhao et~al.(2024)Zhao, Wu, Wang, Tang, Wang, and de~Rijke]{zhao2024toolrec}
Yuyue Zhao, Jiancan Wu, Xiang Wang, Wei Tang, Dingxian Wang, and Maarten de~Rijke.
\newblock Let me do it for you: Towards llm empowered recommendation via tool learning.
\newblock \emph{arXiv preprint arXiv:2405.15114}, 2024.

\bibitem[Zhong et~al.(2025)Zhong, Wang, Ye, Zhang, and Zhu]{zhong2025ggbond}
Hailin Zhong, Hanlin Wang, Yujun Ye, Meiyi Zhang, and Shengxin Zhu.
\newblock Ggbond: Growing graph-based ai-agent society for socially-aware recommender simulation.
\newblock \emph{arXiv preprint arXiv:2505.21154}, 2025.

\bibitem[Zhong et~al.(2024)Zhong, Guo, Gao, Ye, and Wang]{zhong2024memorybank}
Wanjun Zhong, Lianghong Guo, Qiqi Gao, He~Ye, and Yanlin Wang.
\newblock Memorybank: Enhancing large language models with long-term memory.
\newblock In \emph{Proceedings of the AAAI Conference on Artificial Intelligence}, 2024.

\bibitem[Zhou et~al.(2026)Zhou, Zhao, Li, Evmorfos, Demirci, Wang, Liu, Wang, Li, Li, et~al.]{zhou2026llm}
Yuhang Zhou, Zhuokai Zhao, Ke~Li, Spilios Evmorfos, G{\"o}kalp Demirci, Mingyi Wang, Qiao Liu, Qifei Wang, Serena Li, Weiwei Li, et~al.
\newblock Llm-driven reasoning for constraint-aware feature selection in industrial systems.
\newblock \emph{arXiv preprint arXiv:2603.24979}, 2026.

\end{thebibliography}

\clearpage
\newpage
\beginappendix

\section{Sample Prompt}
\label{app:prompt}

To make the harness concrete, Figure~\ref{fig:prompt} gives an abstracted template of the per-cycle decision prompt---the single call in which the model turns the tool-computed context into a proposed configuration. The prompt is abstracted to its operative structure: internal metric, model, and system names are replaced with role placeholders in braces. Both case studies use the same template, differing only in what fills these placeholders (Table~\ref{tab:prompt-instantiation}).

\begin{figure}[H]
\begin{tcolorbox}[colback=gray!3, colframe=gray!50,
  title={\textbf{Per-Cycle Decision Prompt}}, fonttitle=\small,
  fontupper=\small, left=4pt, right=4pt, top=2pt, bottom=2pt]
\begin{verbatim}
System: You are the reasoning core of an agentic harness that
continually optimizes a production recommender. Each cycle you receive
tool-computed context and propose how to reallocate a bounded budget
across the system's control units to improve engagement. A
constrained-optimizer tool then verifies your proposal and, only if it
would exceed the budget, projects it onto the nearest budget-feasible
configuration before deployment.

# Context (produced by the harness's tools)
- Analysis: for each control unit, its current {engagement metrics}
  and {operating cost}, and their change since the previous cycle.
- Budget: the operating budget {B} and each unit's feasible range of
  settings.
- Memory: your configurations from the last {m} cycles and the
  outcomes attributed to them.
- Attribution: the estimated effect of your most recent configuration.

# Task
For each control unit, judge how efficiently its current setting
converts operating budget into engagement, then propose a bounded
adjustment from that unit's feasible set: a continuous multiplier
within a fixed range, or a choice from an ordered discrete menu.
Select only from this set; do not invent actions. Keep the total
within the operating budget {B}. Give a brief rationale and a
confidence for each.

# Output (JSON)
{
  "decisions": [
    {"unit": "<id>", "adjustment": "<action from feasible set>",
     "rationale": "<why>", "confidence": "high|medium|low"}
  ],
  "assessment": "<cross-unit synthesis of what is working and
   what is not>"
}
\end{verbatim}
\end{tcolorbox}
\caption{Abstracted template of the per-cycle decision prompt. Placeholders in braces are filled at runtime by the harness's tools; the model proposes bounded per-unit adjustments, which the constrained optimizer renders budget-feasible before deployment.}
\label{fig:prompt}
\end{figure}

\begin{table}[H]
  \centering
  \caption{How the two case studies instantiate the template's placeholders.}
  \label{tab:prompt-instantiation}
  \begin{tabular}{lll}
    \toprule
    Placeholder & Retrieval-budget case & Serving-capacity case \\
    \midrule
    control unit & retrieval source & user segment \\
    feasible set & bounded continuous multiplier & ordered discrete treatment menu \\
    engagement metric & video-viewing sessions, watch time & engaged sessions, time spent \\
    operating budget & retrieval budget & serving compute budget \\
    \bottomrule
  \end{tabular}
\end{table}

\end{document}